\documentclass[conference]{IEEEtran}
\IEEEoverridecommandlockouts
\usepackage{cite}
\usepackage{amsmath,amssymb,amsfonts}
\usepackage{algorithmic}
\usepackage{algorithm}
\usepackage{graphicx}
\usepackage{textcomp}
\usepackage{xcolor}
\usepackage{subcaption}
\def\BibTeX{{\rm B\kern-.05em{\sc i\kern-.025em b}\kern-.08em
    T\kern-.1667em\lower.7ex\hbox{E}\kern-.125emX}}
\begin{document}

\title{A Real-time Critical-scenario-generation Framework for Testing Autonomous Driving System\\
}

\author{\IEEEauthorblockN{1\textsuperscript{st} Yizhou Xie}
\IEEEauthorblockA{\textit{School of Mechanical Engineering} \\
\textit{Shanghai Jiao Tong University}\\
Shanghai, China \\
xieyizhou@sjtu.edu.cn}
\and
\IEEEauthorblockN{2\textsuperscript{nd} Kunpeng Dai}
\IEEEauthorblockA{\textit{National Engineering Laboratory for Automotive Electronic Control Technology} \\
\textit{Shanghai Intelligent and Connected Vehicle R\&D Center}\\
Shanghai, China \\
dai.kunpeng@icv-ip.com}
\and
\IEEEauthorblockN{3\textsuperscript{rd}Yong Zhang}
\IEEEauthorblockA{\textit{School of Mechanical Engineering} \\
\textit{Shanghai Jiao Tong University}\\
Shanghai, China \\
yongzhang1977@sjtu.edu.cn}
}

\maketitle

\begin{abstract}
In order to find the most likely failure scenarios which may occur under certain given operation domain, critical-scenario-based test is supposed as an effective and widely used method, which gives suggestions for designers to improve the developing algorithm. However, for the state of art, critical-scenario generation approaches commonly utilize random-search or reinforcement learning methods to generate series of scenarios for a specific algorithm, which takes amounts of computing resource for testing a developing target that is always changing, and inapplicable for testing a real-time system. In this paper, we proposed a real-time critical-scenario-generation (RTCSG)  framework to address the above challenges. In our framework, an aggressive-driving algorithm is proposed in controlling the virtual agent vehicles, a specially designed cost function is presented to guide scenarios to evolve towards critical conditions, and a self-adaptive coefficient iteration is designed that enable the approach to operate successfully in different conditions. With our proposed method, the critical-scenarios can be directly generated for the target under test which is a black-box system, and the real-time critical-scenario test can be brought into reality. The simulation results show that our approach is able to obtain more critical scenarios in most conditions than current methods, with a higher stability of success. For a real-time testing, our approach improves the efficiency around 16 times.
\end{abstract}


\section{Introduction}
Amounts of newborn and burgeoning automotive industries has arisen in the past few years, while the complexity of introducing automated driving on a larger scale still poses formidable socio-technical challenges \cite{b1}. For the transformation from artificial driving to autonomous driving, safety validation becomes one of the key challenge, due to the complexity and uncertainty of the driving environments \cite{b3}.

Manufacturers detect and correct defects for conventional vehicles in performance by road testing with mileage accumulation of millions of kilometers, but this tends to be dangerous and time consuming for autonomous vehicles, which are real-time embedded systems with multiple components \cite{b21}.
For the safety validation of autonomous vehicles, typically, simulation test based on scenarios is used to find failures of a system \cite{b2}, which is less expensive than field-testing for evaluating autonomous vehicles \cite{b5,b6,b7}. 
More importantly, simulation test can more easily probe critical scenarios that cannot be obtained in real-world environments \cite{b4}.

Exploring logical scenarios with parameter trajectories \cite{b3}, which are also called
stress testing in \cite{b4,b8,b16} or adversarial testing in \cite{b9}, is different from the scenario reconstructed with real data \cite{b18,b19,b20}. Exploring logical scenarios with parameter trajectories aims at looking for the failure or likely-failure situations of an autonomous system, giving boundary situation or proper suggestions of Operational Design Domain for system designers.
The scenarios are modeled through assumed parameters and uncertain parameters \cite{b3}. Assumed parameters (e.g. realistic driving model, and proper driving rules) ensure the reliability of the scenarios, and uncertain parameters are optimized during generation to find the scenarios of interest.

In \cite{b14,b17}, rapidly exploring random trees (RRT) is used to search for the trajectories of agent vehicles in scenario, that bring Ego into an unsafe state, the trajectory is obtained through randomly generated state, according to the vehicle model and some trajectory generation rules. 
In \cite{b16}, Monte Carlo tree search and reinforcement learning (RL) is used to model the pedestrian-vehicle scenario to extend an adaptive stress testing (AST), which find failures for a black-box system. 
In \cite{b13}, RL-base method combining Signal Temporal Logic as lightweight is used to generate vehicle-vehicle critical scenario, including the driving conditions of car following and cut-in.

Though there already exists some researches in generating scenarios for testing autonomous, current methods still have problems, especially in practicability and efficiency.
RRT-based method requires that the system is able to rewind the generation and restart from certain node, or the scenario have to repeat from the origin for each generation step \cite{b14}, which consumes marvelous time and is infeasible for a real-time system.
RL-based method runs more quickly than a common Monte Carlo tree search in finding interested scenarios \cite{b16}, 
and is able to realize self-evolution during the on-line operation, by receiving rewards through series of generations. However, the time consumption for learning is still at a high level, which commonly take several hours to obtain one interested case \cite{b13,b15,b16}.
In general, current methods for searching an interested scenario are not applicable for real-time tests, and extremely time consuming even for simulation tests.

Further more, AST in \cite{b15,b16} focus on the failure discovery, while put less attention on the form of failure. As pointed out in \cite{b14},
instead of an unavoidable collision, the almost-avoidable collisions or near-misses can give the ability boundary and a more probable form of failure of the target under test.

In this paper, we propose an approach to generate critical scenarios for the target under test which is black-box and operating in real time.
The contributions of this paper can be summarized as follow:
\begin{itemize}
	\item We propose an on-line framework which is able to generate specific scenarios against a black-box target, and we proved that the approach has better performance and higher efficiency than current methods.
	\item We propose a formulated cost function for the aggressive driving behavior, which will lead the scenario to dangerous, while not extreme.
	\item We propose a self-adaptive iteration for coefficients in the cost function, increase the adaptability to different initial conditions, the tests results show that the approach gain a more stable performance in generation.
\end{itemize}

This paper is organized as follows: Section \ref{sec:scenario_def} provides the consideration of assessments, as well as the definitions of generated scenarios. Section \ref{sec:scenario_gen} presents the specific methodology for achieving a critical scenario, including the overall structure, cost function design and coefficients iteration. Section \ref{sec:test} describes the simulation tests and their results, compared with RRT-based and RL-based methods. Finally, Section \ref{sec:conclussion} is the conclusion and future work.

\section{Scenario Definition}\label{sec:scenario_def}

\subsection{Scenario-based Stress Testing}
As described above, testing scenarios for autonomous has different expectations due to the purposes. In this paper, we focus on generating the critical scenario of vehicle-vehicle, that lead to behaviors at the proximity of boundary between collisions and near-collisions \cite{b14}. 
The critical scenario is adaptive to the target under test, and we consider that the assessment criteria is composed with two factors as: 
\begin{itemize}
	\item The gap between agent vehicle's probable state and its ideal state.
	\item Whether the generated scenario conforms to requirements.
\end{itemize}

\begin{figure}[htbp]
	\centering{\includegraphics[width=0.5\textwidth]{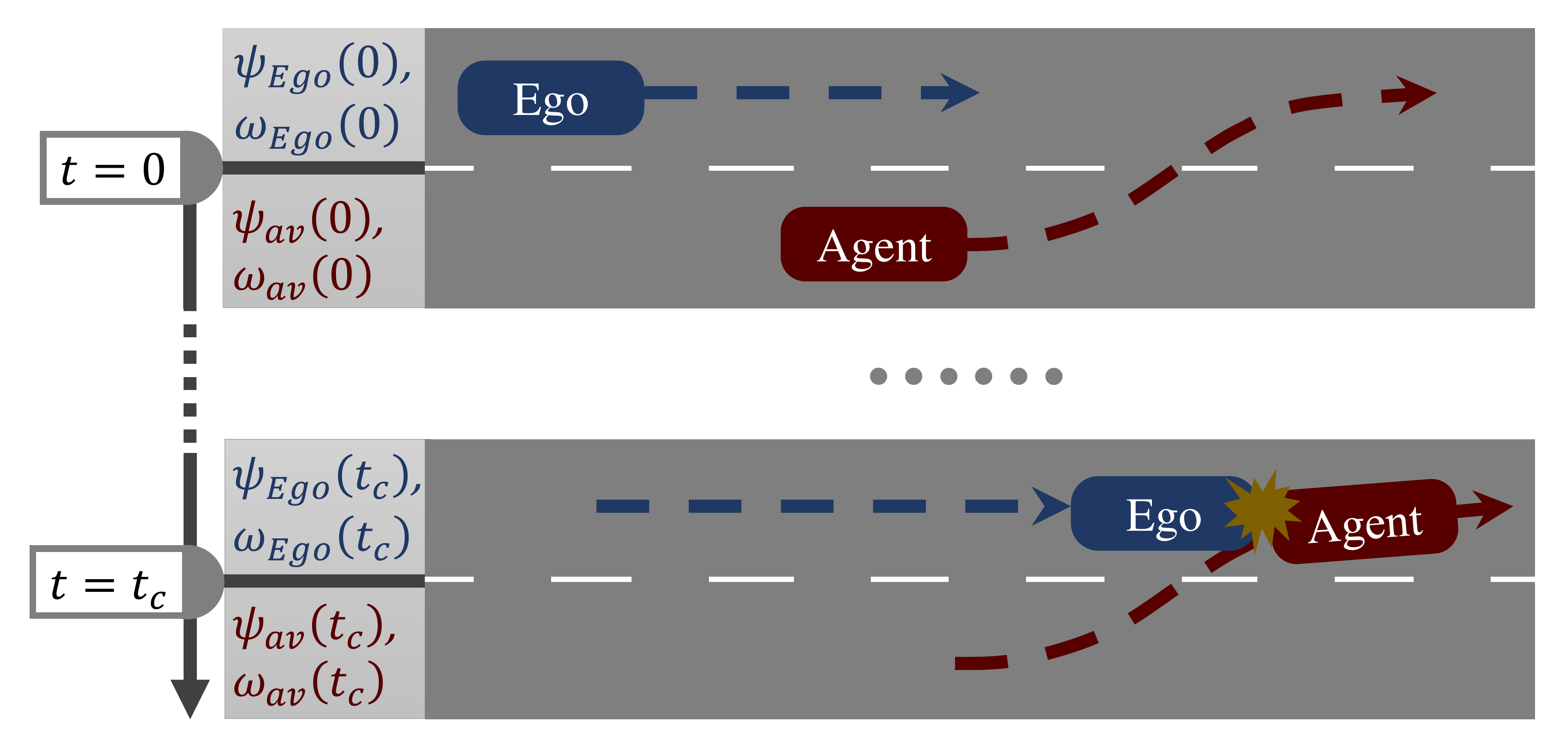}}
	\caption{Diagram of a vehicle-vehicle cut-in scenario.}
	\label{fig:scenario}
\end{figure}

For the first, an ideal state means the agent vehicle form an perfect criticality with Ego (e.g. Ego and agent vehicle collide with the same speed in a car following condition). 
For the second, we expect the performance of agent vehicle accord with the scenario type that may advance specified (e.g. generating the cut-in scenario only). Furthermore, the agent vehicle should threaten while obeying the traffic laws, which avoids the scenario generated being meaningless.

\subsection{Problem Statement}
Scenario state $ \ell $ is described by 
$ \ell=(\Psi,\Omega|\Lambda,t) $, where 
$ \Lambda $ is the set of vehicles (e.g. Ego and agent vehicle), 
$ t $ is the time of scenario, 
$ \Psi $ is the set of vehicle states, 
$ \Omega $ is the input of scenario (i.e. actions of vehicles).
During the generation, a scenario is operated as follow:
\begin{subequations}\label{eq:scenario_def}
	\begin{align}
		\Psi^* &=\Psi+\mathcal{G}(\Omega,\epsilon)\label{eq:scenario_defA}\\
		\Omega^* &=\mathcal{C}(\Psi)\label{eq:scenario_defB}\\
		\ell^* &=(\Psi^*,\Omega^*|\Lambda,t+\epsilon)\label{eq:scenario_defC}
	\end{align}
\end{subequations}
where $ \mathcal{G} $ is the Scenario Transfer (e.g. kinematic/dynamic model, or a real vehicle system, etc.), $ \mathcal{C} $ is the controllers of vehicles, and the scenario $ \ell $ is turned into next step as $ \grave{\ell} $ given a step size $ \epsilon $.

In this paper, we take the cut-in scenario as example, as shown in ``Fig.~\ref{fig:scenario}''.
The states of Ego and agent vehicle at time $ t $ are denoted as
$ \psi_{Ego}(t),\psi_{av}(t)\subseteq\Psi(t) $, 
similarly,
$ \omega_{Ego}(t),\omega_{av}(t)\subseteq\Omega(t) $
denotes the actions of Ego and agent vehicle, separately. 
State $ \psi_\lambda(t) $ of each vehicle $ \lambda\in\Lambda $ is represented by a 4-tuple:
\begin{equation}			
	\psi_\lambda(t)=[X_\lambda(t),Y_\lambda(t),\Theta_\lambda(t),V_\lambda(t)]\label{eq:state_def1}
\end{equation}
where $ X,Y,\Theta,V $ denote the longitudinal position, lateral position, vehicle Yaw and velocity.
In the same way, we define $ \omega_\lambda(t) $ as a 2-tuple:
\begin{equation}			
	\omega_\lambda(t)=[\dot{V}_\lambda(t),\delta_\lambda(t)]\label{eq:state_def2}
\end{equation}
where $ \dot{V} $ is the acceleration and $ \delta_\lambda $ is the steering angle.
In every step of generation, $ \omega_{Ego}(t) $ is obtained from Ego controller, and $ \omega_{av}(t) $ is obtained through our proposed approach, which aims at causing a critical condition.

\begin{figure}[htbp]
	\centering{\includegraphics[width=0.5\textwidth]{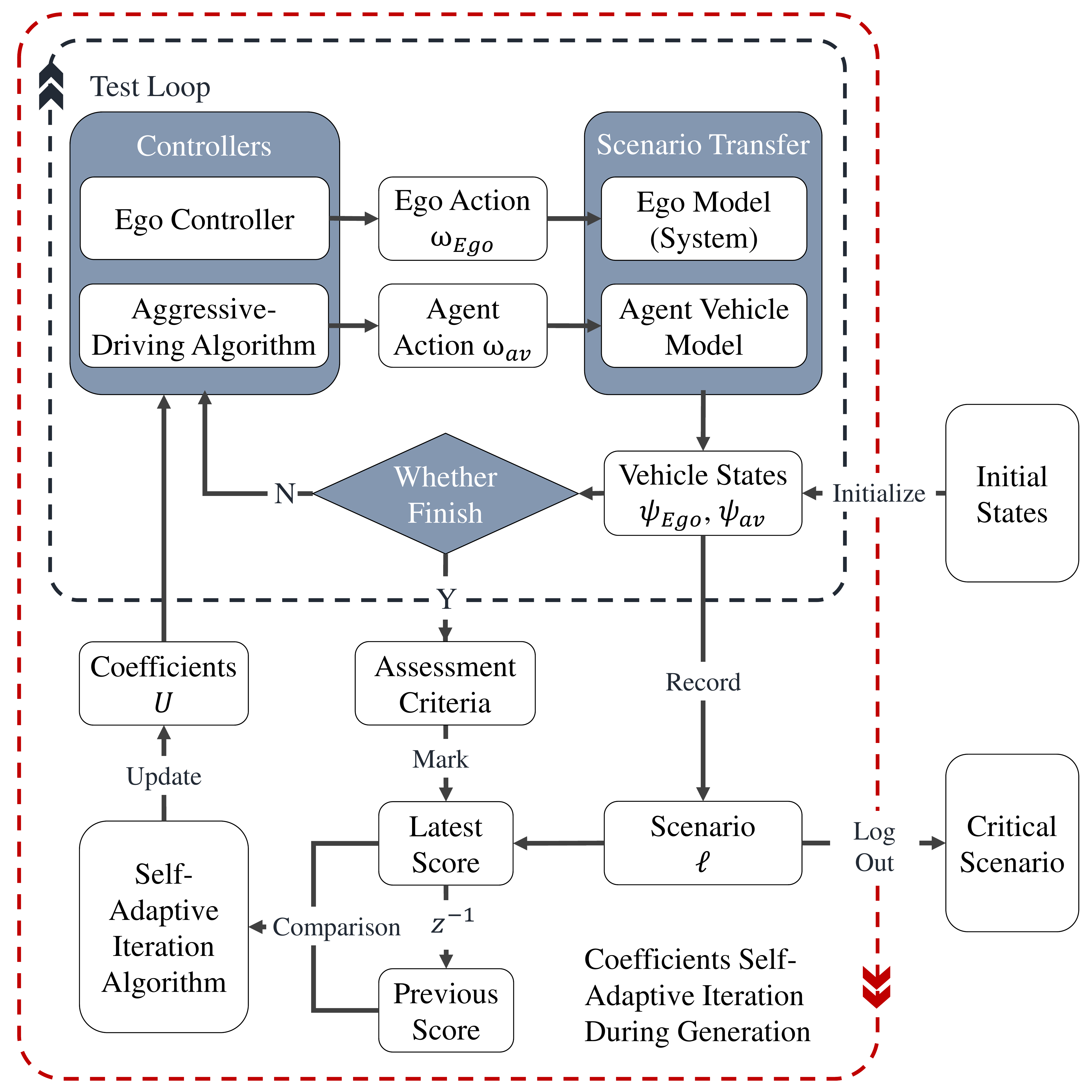}}
	\caption{Structure of RTCSG framework.}
	\label{fig:structure}
\end{figure}

\section{On-line Scenario Generation}\label{sec:scenario_gen}

\subsection{Methodology Framework}

In order to generate a critical scenario in real time, we design an
aggressive-driving algorithm
for the agent vehicle to obtain $ \omega_{av} $, which is similar to the autonomous control algorithm from certain perspectives.
Both of the algorithms aim to avoid vehicle collisions, however, the autonomous algorithm expects to keep a proper safety margin, while the target of 
aggressive-driving algorithm
is to minimize the safety margin.

The general structure of the RTCSG framework is shown as ``Fig.~\ref{fig:structure}''.
During each test loop, system captures the driving states of Ego and agent vehicle, then the
aggressive-driving algorithm
which is designed based on MPC method is operated and the best action of agent vehicle is chosen,
and the agent vehicle will be lead to approaching Ego while without causing unavoidable collision.
Note that the model of Ego in Scenario Transfer can be substituted by actual vehicle hardware system during a Vehicle in the Loop (VIL) or Hardware in the Loop (HIL) test, and the hardware system of Ego will also be the part under test.
For MPC, the coefficients of cost function will significantly impact the control effects, and it is hard to find a universal coefficient series for all different conditions.
In this system, during each generation, a score will be obtained to appraise the criticality of generated scenario, and the coefficients will iterate after each generation according to the latest score and the previous one.

\subsection{Predicting Model}
The kinematic-based predicting model (denoted as $ \mathcal{K} $) proposed in \cite{b11} is utilized to estimate vehicle states $ \tilde{\psi}_{Ego}(\tau|t) $ and $ \tilde{\psi}_{av}^\zeta(\tau|t) $ in the near future:
\begin{subequations}\label{eq:predict_mdl}
	\begin{align}
		\tilde{\psi}_{Ego}(\tau|t)&=\psi_{Ego}(t)+\mathcal{K}[\psi_{Ego}(t),\omega_{Ego}(t),\tau]\label{eq:predict_mdla}\\	\tilde{\psi}_{av}^\zeta(\tau|t)&=\psi_{av}(t)+\mathcal{K}[\psi_{av}(t),\zeta,\tau]\label{eq:predict mdlb}
	\end{align}
\end{subequations}
where $ \tilde{\psi}_{Ego}(\tau|t) $ denotes the predicting state of Ego after $ \tau $ seconds given the current state at time $ t $, and action $ \omega_{Ego}(t) $ is the output from Ego controller.
Similarly, $ \tilde{\psi}_{av}^\zeta(\tau|t) $ is the predicting state of agent vehicle, and 
$ \zeta\subset\omega_{av} $ 
denotes the candidate action that agent vehicle plans to take.

This kinematics-based prediction is applied for every step during the generation, note that the framework we proposed allows select other models to do the prediction (e.g. prediction based on polynomial curve, dynamic model, neural network, etc.), which depends on the target under test and environment configuration.

\subsection{Cost Function Design}
After the state prediction 
$ \tilde{\psi}_{Ego}(\tau|t) $, 
$ \tilde{\psi}_{av}^\zeta(\tau|t) $ 
of each action given the current state 
at time $ t $ is calculated, we utilize a cost function to evaluate the criticality.
We define the ideal target state for the agent vehicle in scenario at time $ t $ as:
\begin{equation}
	\psi_{idl}(t)= \psi_{Ego}(t)+\psi_{f}\label{eq:idl_state1}
\end{equation}
and similarly, the ideal target state given a predicting period $ \tau $ as:
\begin{equation}
	\tilde{\psi}_{idl}(\tau|t)= \tilde{\psi}_{Ego}(\tau|t)+\psi_{f}\label{eq:idl_state2}
\end{equation}
where $ \psi_{f} $ is the states offset between Ego and agent vehicle in an ideal critical situation.

In an ideal critical cut-in condition, Ego is expected to have a tailgating with the front agent vehicle, while owning the same state except the longitudinal positions, so in this paper, we present
$ \psi_{f}=[d,0,0,0] $ 
in a cut-in scenario, where 
$ d=(l_{Ego}+l_{av})/2 $ 
is the longitudinal position distance when two vehicles collide,
$ l_{Ego} $ and $ l_{av} $ are the length of two vehicles.
The cost function proposed for criticality computing is defined as follow:
\begin{subequations}\label{eq:cost_fun}
	\begin{align}
	    \mathcal{J}_1^\zeta(\tau|t)&=\mathcal{D}^\zeta(\tau|t)\ \mathcal{U}\ [\mathcal{D}^\zeta(\tau|t)]^T \label{eq:cost_funA}\\	\mathcal{D}^\zeta(\tau|t)&=\tilde{\psi}_{av}^\zeta(\tau|t)-\tilde{\psi}_{idl}(\tau|t)\label{eq:cost_funB}\\
		\mathcal{U}&=\mathrm{diag}\  U\label{eq:cost_funC}
	\end{align}
\end{subequations}
where $ U=[u_1, u_2, \cdots, u_m] $ are the coefficients corresponding to the vehicle state $ \psi $, i.e., $ m=4 $ in this paper.

The cost function for keeping the scenario generated conform to requirements is defined as:
\begin{equation}\label{eq:cost_funD}
	{\mathcal{J}_2^\zeta}(\tau|t) =
	\begin{cases}
		\kappa,&{\text{if}}\ \tilde{\psi}_{av}^\zeta(\tau|t)\in E\\
		0,&{\text{otherwise}}
	\end{cases}
\end{equation}
where $ \kappa $ is a positive constant, $ E $ is the state set of agent vehicle that is failed to meet the requests (e.g. unable to merge with Ego in generating cut-in-type scenarios, or violating the traffic laws). Finally, for every step at time $ t $ and given a predicting period $ \tau $, the cost function will lead the agent vehicle presenting the behavior we interested by selecting the best action as:
\begin{equation}\label{eq:act select}
	\omega_{av}(t)=\mathop{\arg\min}_{\zeta}\ [\mathop{\max}_{\tau}\ \mathcal{J}_1^\zeta(\tau|t)+\mathcal{J}_2^\zeta(\tau|t)]
\end{equation}

It should be point out that, in our framework, the cost function above is allowed to be substituted by other designed ones, according to the key focus of the test.

\subsection{Coefficient Self-adaptive Iteration}
Using constant coefficients for the cost function is difficult to achieve ideal result in all different initial states, and we present an self-adaptive iteration for the coefficients.

For each update in iteration, a scenario is generated with the cost coefficients 
$ U=[u_1, u_2, \cdots, u_m] $,
and state $ \ell_i $ of the $ i^{th} $ generated scenario will be logged.
We define $ \mathcal{M}^i(t|L) $ as the Mahalanobis Distance between $ \psi_{idl}(t) $ and $ \psi_{av}(t) $ in $ \ell_i $, which is calculated given $ L=\{\ell_{i_1},\ell_{i_2},\cdots\} $ as the whole sample, 
and the score $ S_i^L $ we use for accepting an update of $ U $ can be shown as:
\begin{subequations}\label{eq:score}
	\begin{align}
		S_i^L&=\dfrac{V^i_{Ego}(t_c^i)}{V^i_{Ego}(t_c^i)+\mathcal{M}^i(t|L)}\label{eq:score1}\\
		t_c^i&=\mathop{\arg\min}_{t}\ \dfrac{V^i_{Ego}(t)+\mathcal{M}^i(t|\ell_i)}{V^i_{Ego}(t)}\label{eq:score2}
	\end{align}
\end{subequations}
where $ V^i_{Ego}(t) $ denotes the velocity of Ego at time $ t $ in the $ i^{th} $ scenario, and $ t_c^i $ is the most critical moment chosen during each generation.
As shown in (\ref{eq:score}), we consider that less difference from the ideal with higher driving speed will lead to a more critical condition. 

\begin{algorithm}[H]
	\caption{Self-adaptive Iteration of Coefficients in the Cost Function.}\label{alg:alg_Coef_iter}
	\begin{algorithmic}
		\STATE 
		\STATE {\textsc{Coef}}$(R,  n_{prev})$	
		\STATE \hspace{0.5cm}\textbf{if} $ \exists n_{prev} $ \textbf{then}
		\STATE \hspace{1.0cm}\textbf{if} $ R>1 $ \textbf{then}
		\STATE \hspace{1.5cm}$ \Gamma(n_{prev})\gets  \Gamma(n_{prev})*\beta $
		\STATE \hspace{1.0cm}\textbf{else}
		\STATE \hspace{1.5cm}$ \Gamma(n_{prev})\gets  \Gamma(n_{prev})/\beta  $
		\STATE \hspace{1.0cm}\textbf{end if}	
		\STATE \hspace{1.0cm}\textbf{if} $ rand(0,1)<e^{(R-1)/(K*T)} $ \textbf{then}
		\STATE \hspace{1.5cm}$ T\gets  T/\alpha $
		\STATE \hspace{1.5cm}$ NumberOfFails=0  $	
		\STATE \hspace{1.0cm}\textbf{else}
		\STATE \hspace{1.5cm}$ NumberOfFails=NumberOfFails+1 $
		\STATE \hspace{1.5cm}$ U(n_{prev})\gets U(n_{prev})/[1+\Gamma(n_{prev})*\beta] $
		\STATE \hspace{1.0cm}\textbf{end if}	
		\STATE \hspace{0.5cm}\textbf{end if}
		\STATE \hspace{0.5cm}\textbf{if} $ NumberOfFails>maxNumberOfFails $	
		\STATE \hspace{1.0cm}\textbf{break}
		\STATE \hspace{0.5cm}\textbf{end if}
		\STATE \hspace{0.5cm}$ \textbf{select randomly } n \in [1, 2, \cdots, m] $
		\STATE \hspace{0.5cm}$ U(n)\gets U(n)*[1+\Gamma(n)] $
		\STATE \hspace{0.5cm}$ n_{prev}\gets n $
		\STATE \hspace{0.5cm}\textbf{return} $ U, \Gamma $
	\end{algorithmic}
	\label{alg_Coef_iter}
\end{algorithm}

In order to employ Mahalanobis Distance as the assessment criteria, scenarios will be scored in pairs to keep the Mahalanobis Distance values in the same dimension.
For $ i^{th} $ generated scenario, 
we define $ R $ as the ratio of scores between the latest scenario generated and its previous one:
\begin{equation}
	R=\dfrac{S_{i}^{\{\ell_i,\ell_{i-1}\}}}{S_{i-1}^{\{\ell_i,\ell_{i-1}\}}}
\end{equation}

The iteration is shown as ``Alg.~\ref{alg:alg_Coef_iter}'', we adopt a temperature-based judgment in iteration which is similar to the one proposed in\cite{b10},
where $ \alpha $  and $ K $ are constant parameters, and $ T $ is the temperature parameter controlling the likelihood of acceptance. $ \Gamma=[\gamma_1,\gamma_2,\cdots,\gamma_m] $ is the increasing rate corresponding to $ U $, and $ \beta $ is a constant which governing the gradient of $ U $, which ensure the iteration convergent after the coefficients achieving proper values.

\section{Test and Result}\label{sec:test}
To demonstrate the ability of proposed approach, we constructed a case study shown in ``Fig.~\ref{fig:scenario}''.
In our tests, the agent vehicle is requested to do the cut-in during a lane change maneuver, given different initial states.

We denote $ \Delta X=X_{av}(0)-X_{Ego}(0) $ as the initial difference of longitudinal positions between Ego and agent vehicle, and denote $ \Delta V=V_{av}(0)-V_{Ego}(0) $ as the initial difference of velocities.

Initial states of Ego $ \psi_{Ego}(0) $ in the tests are fixed, instead, the longitudinal position $ X_{av}(0) $ and initial speed $ V_{av}(0) $ of agent vehicle is changed for different scenarios. In the beginning of a scenario, the difference of lateral positions $ \left|Y_{Ego}(0)-Y_{av}(0)\right| $ is set as the lane width, and $ \Theta_\lambda(0) $ are all zeros at start.

Target under test (i.e. control algorithm of Ego) is modeled according to an adaptive cruise controller proposed in \cite{b12}, and the control algorithm is enabled when the lateral distance between two vehicles less than a default value.

\subsection{Scenario Generated}
\begin{figure}[htbp]
	\centering
	\begin{subfigure}[b]{0.5\textwidth}
		\centering
		\includegraphics[width=\textwidth]{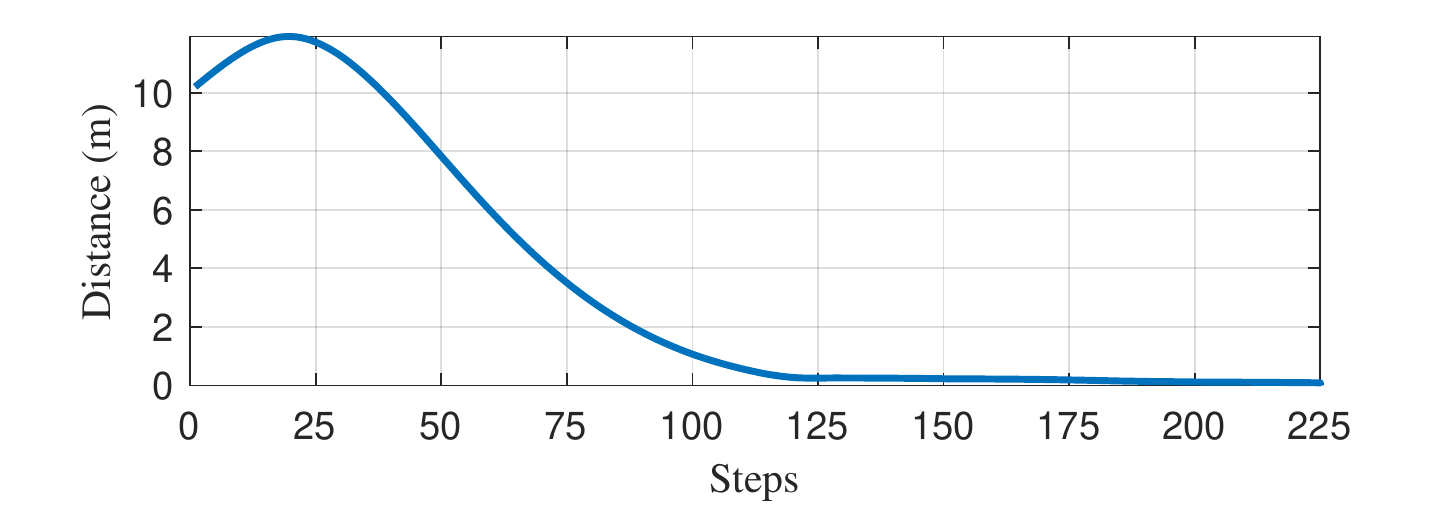}
		\caption{Distance between Ego and agent vehicle}
		\label{fig:scenario_example_x}
	\end{subfigure}
	\begin{subfigure}[b]{0.5\textwidth}
		\centering
		\includegraphics[width=\textwidth]{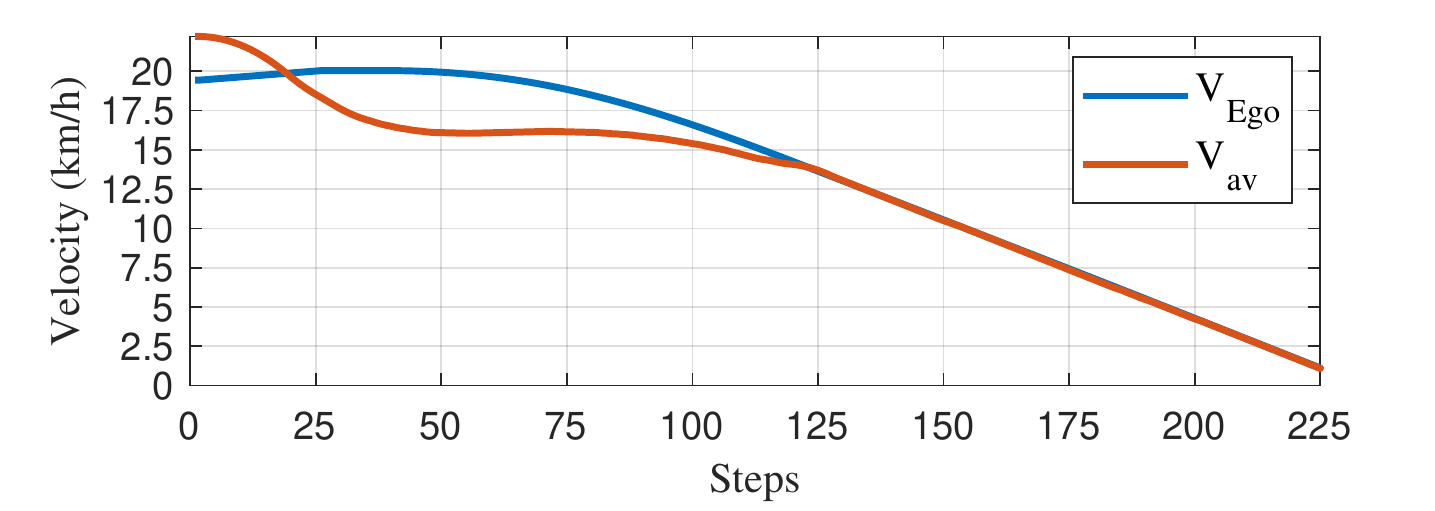}
		\caption{Velocity episode of Ego and agent vehicle}
		\label{fig:scenario_example_vel}
	\end{subfigure}
	\caption{Generated critical scenario given $ \Delta X=15 $ (m) and $ \Delta V=10 $ (km/h).}
	\label{fig:scenario_example}
\end{figure}

We carry out the tests based on MATLAB, and ``Fig.~\ref{fig:scenario_example}'' gives one of the scenarios generated through the proposed approach. In this scenario, initial longitudinal position of agent vehicle is 15 meters front of Ego, the initial driving speed of Ego is 70 km/h, and the one of agent vehicle is 80 km/h.

Episodes in ``Fig.~\ref{fig:scenario_example}'' shows that agent vehicle is able to threaten Ego and induce a critical condition.
``Fig.~\ref{fig:scenario_example_x}'' displays the distance between Ego and agent vehicle, for the length of two vehicles are all set to 4.8 meters in the tests, the actual distance between two vehicles is 10.2 meters at beginning.
At the moment that collision nearly happen (i.e. distance is reduced close to 0, at the $ 122^{th} $ step), the velocities of two vehicles becomes almost the same, as shown in ``Fig.~\ref{fig:scenario_example_vel}'', which suggests the accident is not unavoidable. It should be pointed out that the scenario is generated in a real time process, the controller of Ego is a black box to the generation system, and a satisfactory scenario is obtained just during several times of running.

To be specific, as the example shown above, RTCSG achieve a usable critical scenario in 470 steps (each step is set to 0.05 second in the tests), which took only 23.5 seconds to run. And the approach continue improving scenario score in its following generation, as described in ``Alg.~\ref{alg:alg_Coef_iter}''.

\subsection{Assessment Criteria and Results Comparison}
In this section, we compare the scenarios obtained from our proposed approach with the one from RL-based \cite{b13} and RRT-based methods \cite{b14}.

The way of assessment is same to the score compute in (\ref{eq:score}). For every initial state, scenarios are generated through three different methods, and the most critical moments $ t_c^i $ from scenario $ \ell_i $ are chosen, the score 
$ S_i^{L_{tol}} $
of each method given different initial states can then be obtained, where $ L_{tol} $ is the total set of all $ \ell_i $.

\begin{figure}[htbp]
	\centering
	\begin{subfigure}[b]{0.5\textwidth}
		\centering
		\includegraphics[width=\textwidth]{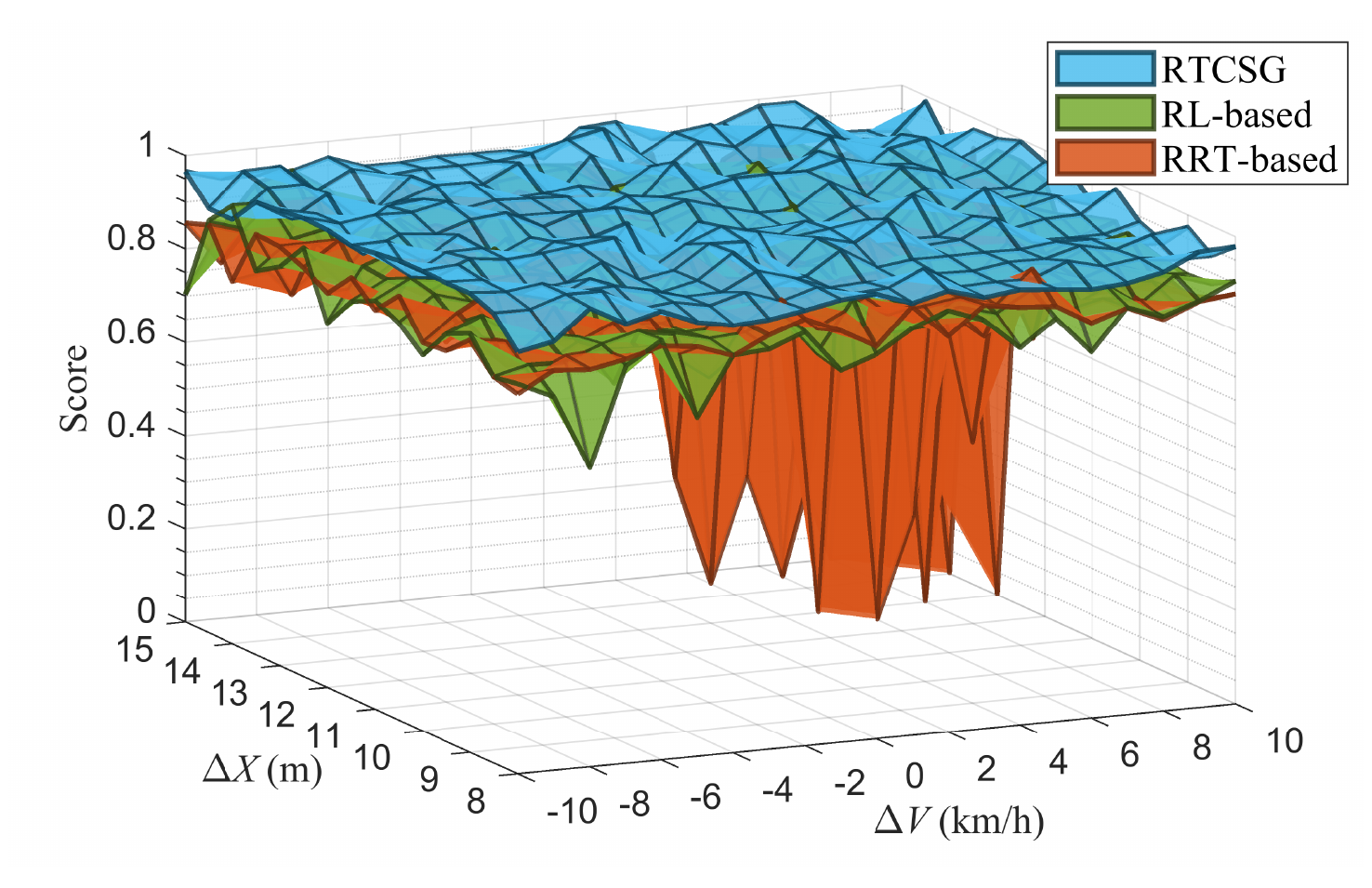}
		\caption{Distribution of the scores given series of initial states}
		\label{fig:score_distribute1}
	\end{subfigure}
	\begin{subfigure}[b]{0.5\textwidth}
		\centering
		\includegraphics[width=\textwidth]{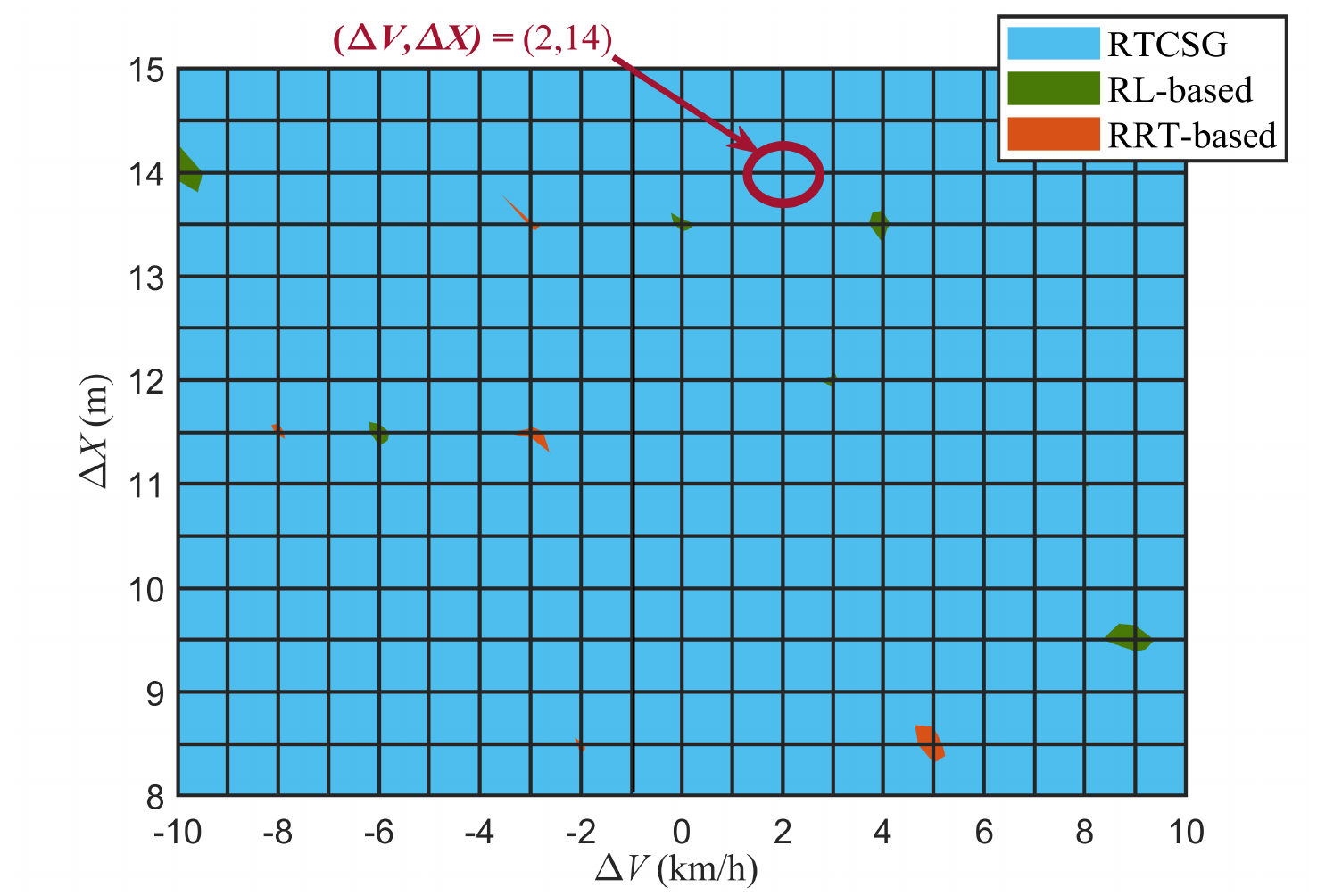}
		\caption{Grid plot of scores given series of initial states}
		\label{fig:score_distribute2}
	\end{subfigure}
	\caption{Scores obtained from three methods given series of initial states.}
\end{figure}

We applied three different methods to generate the scenarios given sets of initial conditions. In the tests, initial velocity of Ego is fixed as 70 km/h, and the range of $ \Delta V (km/h)$ is $ [-10,10] $, meanwhile, the range of $ \Delta X (m)$ is $ [8,15] $. For each initial condition, we do the Monte Carlo run for 5 times, and compute the average result to reduce the impact of randoms in algorithms.

As shown in ``Fig.~\ref{fig:score_distribute1}'', RTCSG has all desired results given different initial conditions, while the other two methods performance unstably when the velocities and positions of two vehicles have a larger initial difference.
For RL and RRT both depend on the feedback of result in every step, which will descend into a failure easily if the initial condition is adverse (e.g. hard to find a proper route direction from start), and the interaction between vehicles is a rapid developing case, which is hard for the searching algorithm to go back a proper track after once deviation.
Colors in ``Fig.~\ref{fig:score_distribute2}'' denote the method which gets the highest score in its covered initial condition area, e.g., the color in grid plot is blue at point $ (\Delta V,\Delta X)=(2,14) $, donates our approach achieves the best result given the initial condition that $ \Delta V=2 $ (km/h) and $ \Delta X=14 $ (m).

In RL-based and RRT-based methods, some stochastic processes are utilized in searching interested events, while in the most part of a critical scenario, the behaviors of agent vehicle are roughly regular to meet certain requirements of the test, and such stochastic processes is unstable and expensive in generating such ruled scenarios.
In our approach, aggressive-driving algorithm makes proper decisions based on designed cost function and coefficient iteration gears the behavior of agent vehicle to a better case.

\begin{figure}[htbp]
	\centering{\includegraphics[width=0.5\textwidth]{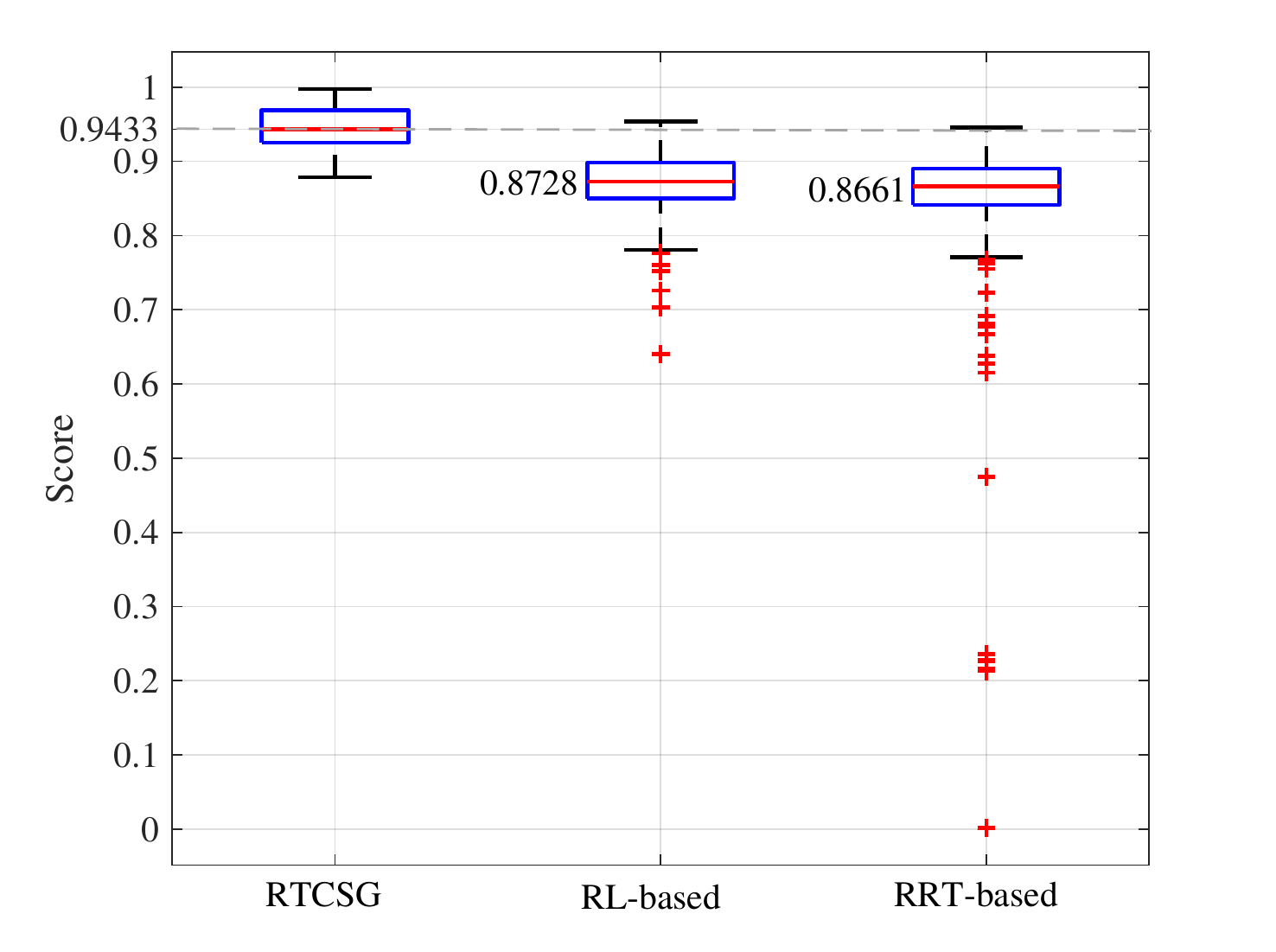}}
	\caption{Box plot of the scores obtained from three methods.}
	\label{fig:score_box}
\end{figure}

As shown in ``Fig.~\ref{fig:score_distribute2}'', for most of the conditions, our approach has a better result,
and ``Fig.~\ref{fig:score_box}'' provides the box plot of the scores achieved by three methods.
Our approach receives smaller mild outliers and no extreme outliers in this test, and obtains the median score at 
0.9433,
higher than the scores obtained from RL-based and RRT-based methods, which are 
0.8728 and 0.8661,
separately.

\subsection{Availability for a Real-time System}
We count the steps spent of each scenario, and the step length in different methods are set to the same fixed size, so that we can assess the time consumption of a scenario generation, which reflects the application costs of methods.

\begin{figure}[htbp]
	\centering
	\begin{subfigure}[b]{0.5\textwidth}
		\centering
		\includegraphics[width=\textwidth]{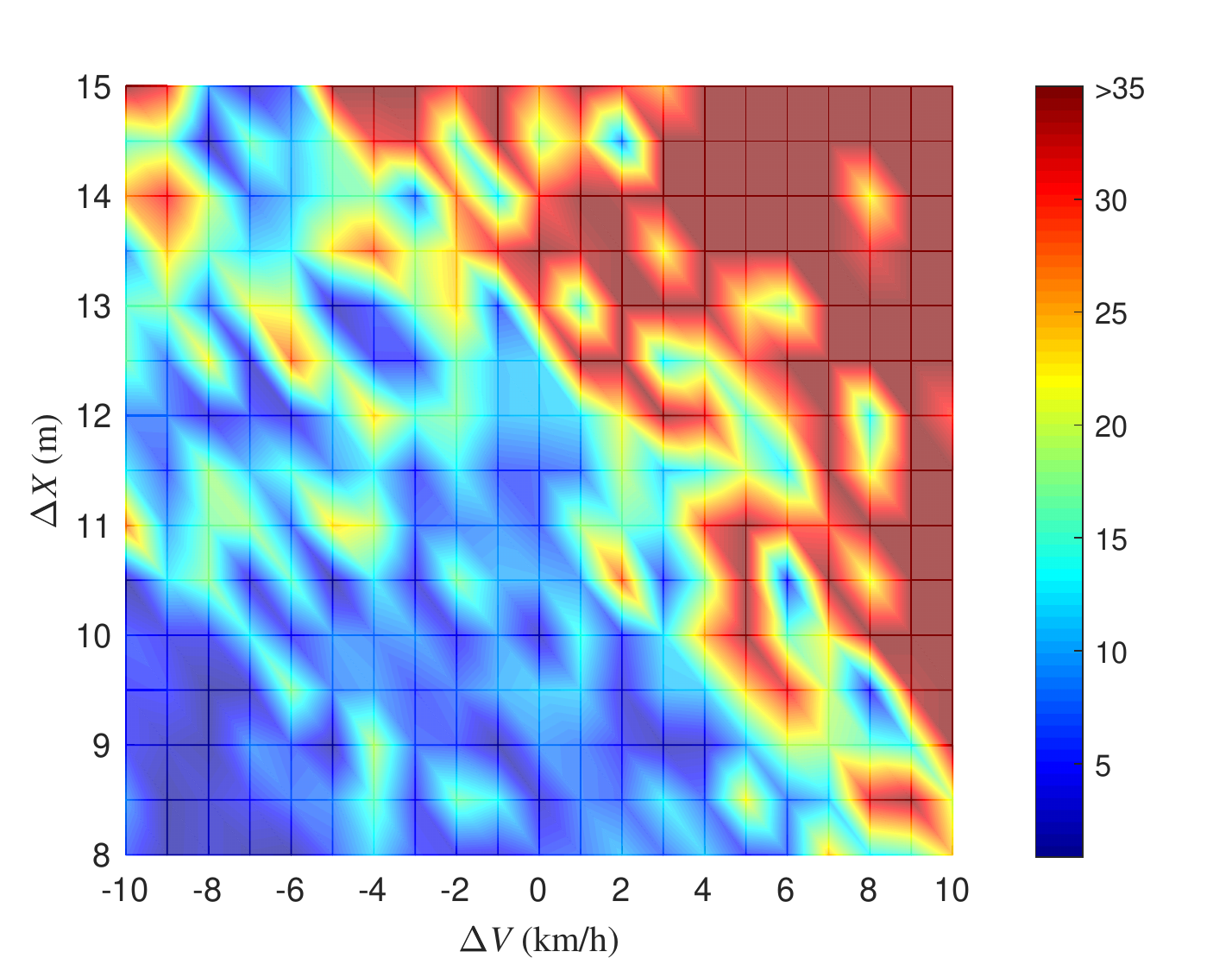}
		\caption{Distribution of the step-spent-ratio given series of initial states}
		\label{fig:ssr_distribute}
	\end{subfigure}
	\begin{subfigure}[b]{0.5\textwidth}
		\centering
		\includegraphics[width=\textwidth]{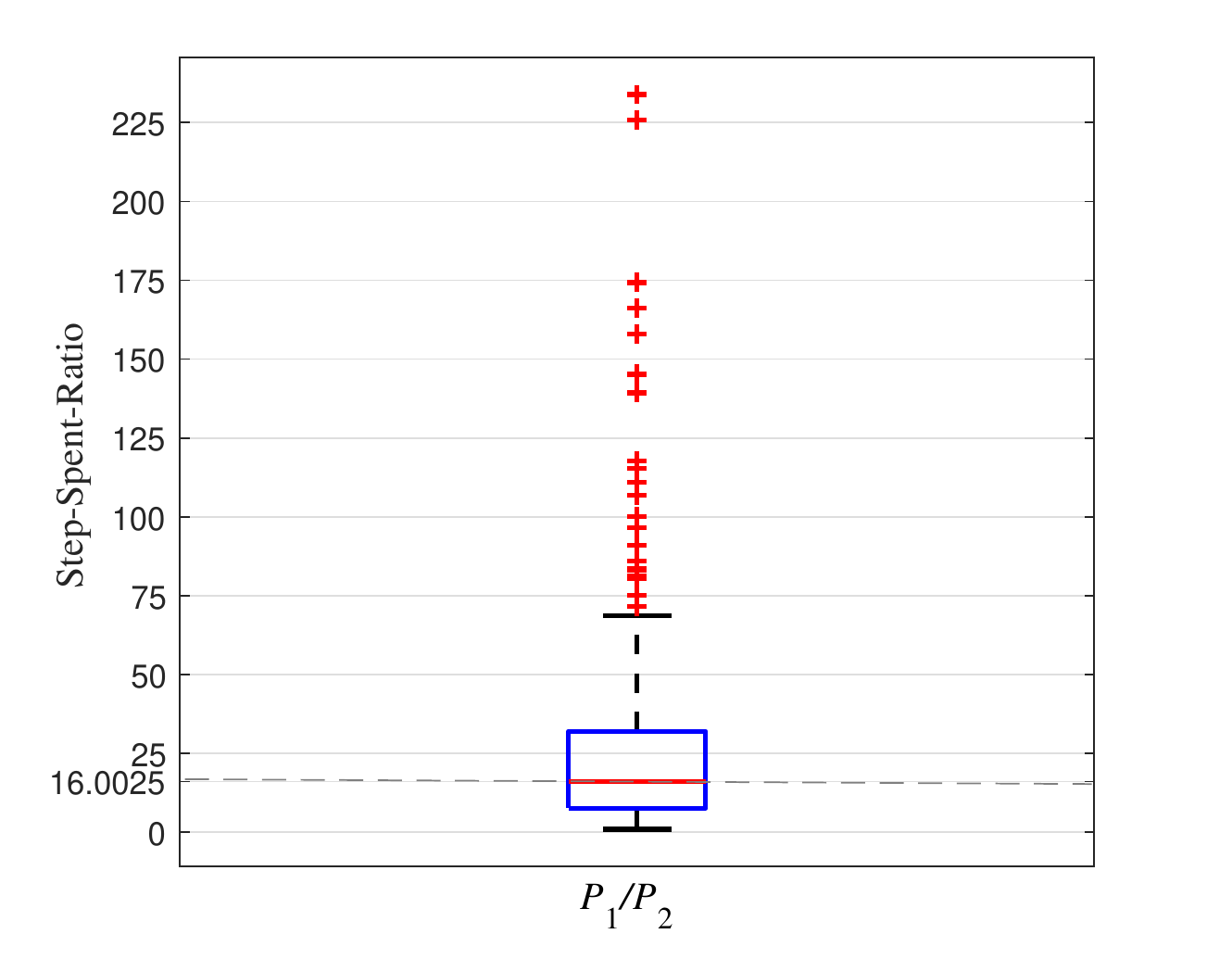}
		\caption{Box plot of step-spent-ratio}
		\label{fig:ssr_box}
	\end{subfigure}
	\caption{Step-spent-ratio between RL-based method and RTCSG method.}
\end{figure}

For the RRT-based method described in \cite{b14} is not suitable in real time generation, in this section, we just compare the time consumption between RTCSG method and the RL-based method.

In the whole generation given a initial state, we record the step numbers till the scenario finally finishes. 
For RL-based method, generation will be stopped till the scenario attain its finish condition, and total steps spent will then be recorded as $ P_{1} $. 
For RTCSG method, generation will be supposed as finished when the coefficient iteration converges, and then we log the steps spent $ P_{2} $.
Finally, the specific value $ P_{1}/P_{2} $ of step numbers is obtained as the step-spend-ratio, and higher $ P_{1}/P_{2} $ denotes more efficient PTCSG performances than the RL-based method.

In ``Fig.~\ref{fig:ssr_distribute}'', we plot the step-spend-ratio of scenario generations given different initial conditions. For RL-based method, algorithm will keep exploring till a satisfactory result is obtained, the ratio appears a sharp ascent in the area that initial state differences of two vehicles are large, where the RL-based method also obtain worse results compared with RTCSG. 
``Fig.~\ref{fig:ssr_box}'' is the box plot of the step-spend-ratio, where the median value indicates that RTCSG approach is able to promote efficiency of scenario generation by 16.0025 times in median.

\section{Conclusion}\label{sec:conclussion}
This paper propose a RTCSG framework to generate critical scenario for a real-time black-box target.
We analysis the requirements of scenario-based test system for black-box autonomous controller, and show the limitation of current scenario generation methods in testing a real-time black-box target.
To address the challenge, we construct an on-line generation framework composed with aggressive-driving algorithm and self-adaptive coefficient iteration. 
A designed cost function formulates the dangerous trajectory, and lead the agent vehicle threaten Ego obeying certain presupposed script, and avoid extreme behaviors.
Our approach is able to generate the scenario against a black-box controller directly, and improve the performance in several following generations, which is more stable and efficient compared with RL-based and RRT-based methods.
The test results show that our approach still achieves expected result given unfavorable initial condition, and performances best in the vast majority of cases, the efficiency of our approach is around 16 times as the RL-based methods in general.

Future work will involve in the incorporation of aggressive-driving algorithm with on-line searching methods, to improve the ability for discovering potential defects of autonomous controller.

\end{document}